\documentclass[sigconf]{acmart}

\usepackage{booktabs} 

\newcommand\rescal{{\scshape Rescal}~}
\usepackage{amsmath}
\usepackage{amsthm}
\usepackage{amssymb}
\usepackage[boxed,lined,algoruled]{algorithm2e}
\usepackage{url}
\usepackage{mathtools}
\usepackage{multirow}
\usepackage{verbatim}
\usepackage{color}
\usepackage{graphicx}
\usepackage{caption}
\usepackage{subcaption}

\usepackage{rotating}
\usepackage{balance}

\setcopyright{none}
\acmConference[KBCOM'18]{Workshop on Knowledge Base Construction, Reasoning and Mining}{Feb 2018}{Los
  Angeles, California USA} 
\acmYear{2018}
\copyrightyear{2018}

\begin{document}
\title{Analysis of the Impact of Negative Sampling \\ on Link Prediction in Knowledge Graphs}

\author{Bhushan Kotnis and Vivi Nastase}
\affiliation{
  \institution{Institute for Computational Linguistics, \\ University of Heidelberg}
  \city{Heidelberg, Germany} \\
\{kotnis,nastase\}@cl.uni-heidelberg.de
}

\begin{abstract}
Knowledge graphs are large, useful, but incomplete knowledge repositories. They encode knowledge through entities and relations which define each other through the connective structure of the graph. This has inspired methods for the joint embedding of entities and relations in continuous low-dimensional vector spaces, that can be used to induce new edges in the graph, i.e., link prediction in knowledge graphs. Learning these representations relies on contrasting positive instances with negative ones. Knowledge graphs include only positive relation instances, leaving the door open for a variety of methods for selecting negative examples. We present an empirical study on the impact of negative sampling on the learned embeddings, assessed through the task of link prediction. We use state-of-the-art knowledge graph embedding methods -- \rescal , TransE, DistMult and ComplEX -- and evaluate on benchmark datasets -- FB15k and WN18. We compare well known methods for negative sampling and propose two new embedding based sampling methods. We note a marked difference in the impact of these sampling methods on the two datasets, with the "traditional" corrupting positives method leading to best results on WN18, while embedding based methods benefit FB15k. 
\end{abstract}




%
%

\maketitle
\section{Introduction}

Much of human knowledge can be formalized in terms of real world entities, abstract concepts, categories and the relations between them. A graph structure -- a knowledge graph (KG) -- is a natural candidate for representing this. NELL \cite{Carlson2010T}, Freebase \cite{Bollacker2008} and YAGO \cite{Suchanek2007} are examples of large knowledge graphs that contain millions of entities and facts. Facts are represented as triples, each consisting of two entities connected by a binary relation, e.g., (\textit{concept:city:London}, \textit{relation:country\_capital}, \textit{concept:country:UK}). Here entities such as London and UK are represented as nodes and the relation \textit{country\_capital} is represented as a binary link that connects these nodes. The same two nodes may be connected by more than one type of relation, making the KG a multi-graph. KGs have found applications in question answering systems \cite{Miller2016}, evaluating trustworthiness of web content \cite{Dong2015}, and web search \cite{Dong2014}. 

Although KGs such as Freebase consist of millions of entities and billions of facts, they are still incomplete \cite{West2014} which limits their application. However, it is possible to infer new (missing) facts from known facts. Recently, latent factor models that capture global patterns from the KG have received considerable attention. They learn a representation of the graph in a continuous vector space by inducing embeddings that capture the graph structure. 

Predicting new edges to automatically add new facts to a KG helps bypass the text analysis stage and bootstrap new knowledge based on what is already captured in the KG. Similar to other problems in processing natural language, such as parsing, data consists (almost) exclusively of positive instances. A solution to this issue is using {\it implicit negative evidence}, whereby instances that have not been observed are considered negatives, and are used for {\it contrastive estimation} \cite{smith2005}, where the aim is to rank observed instances higher than negative (unobserved) ones. Negative instances can be generated using a variety of methods. 

In this article we present the results of our investigation on the impact of several negative sampling methods on state-of-the-art knowledge graph embedding models. Additionally we propose two negative sampling strategies for fine tuning the model. Understanding the impact of negative instance sampling will have at least two uses: providing the basis for choosing the negative sampling method to build the best model for a given method, and allowing us to place in the right context results reported in the literature that were produced while using different negative sampling methods.

\section{Link prediction in Knowledge Graphs}

Knowledge graphs $KG = (\mathcal{E},\mathcal{R})$ contain knowledge in the form of relation triples $(s,r,t)$, where $s,t \in \mathcal{E}$ are entities, and $r \in \mathcal{R}$ is a relation. These knowledge graphs are not complete, and additional links (facts) can be inferred, based on the idea that similar nodes have similar relations -- e.g. all countries have a capital city. 

The KG can be encoded using different modeling techniques, which results in encodings for both the entities and the relations. A variety of techniques have been proposed \cite{Nickel2011,Socher2013,Bordes2013,Yang2015,Lin2015,Nickel2016}. These methods learn a model for the processed KG as a large set of parameters, induced based on optimizing a loss function with respect to positive and negative instances of links representing different relations. Methods such as \rescal \cite{Nickel2011} and Neural Tensor Networks \cite{Socher2013} learn millions of parameters that makes them more flexible, enabling them to model well a variety of relations, but at the cost of increased computational complexity and potential overfitting. TransE \cite{Bordes2013}, DistMult \cite{Yang2015} learn simpler models (with far fewer parameters) and are easier to train but are unable to model certain types of relations such as many-to-one (TransE) and asymmetric relations (DistMult). Recent work such as \cite{Nickel2016} achieve the modeling power of \rescal  with a smaller number of parameters by compressing the tensor product. Complex valued embeddings (ComplEx) \cite{Trouillon2017} extend the DistMult to model antisymmetric relations by using complex valued embeddings. 

\cite{Guu2015} showed that most latent factor models can be modified to learn from paths rather than individual triples which improves performance. Recurrent Neural Networks that learn path representations have also been used for link prediction \cite{Neelakantan2015,Das2016}. All these models require negative samples during training. 

We focus our analysis on four state-of-the-art methods with respect to link prediction in knowledge graphs: ComplEx, DistMult, \rescal, TransE.  ComplEx performs as well as the Holographic Embedding (HolE) model, so HolE was not included\footnote{And also because HolE is very similar to ComplEx. This was verified through personal correspondence with an author of the ComplEx paper.}.

\subsection{\rescal}

The \rescal model \cite{Nickel2011,Nickel2012} weighs the interaction of all pairwise latent factors between the source and target entity for predicting a relation. It represents every entity as a $d$ dimensional vector ($x \in \mathbb{R}^d$), and every relation as a $d\times d$ matrix $W \in \mathbb{R}^{d\times d}$. This model represents the triple $(s,r,t)$ as a score given by
\begin{align*} \label{eqn:loss}
s_c(s,r,t) = x_s^T \ W_r \ x_t
\end{align*}
These vectors and matrices are learned using a loss function that contrasts the score of a correct triple to incorrect ones. Commonly used loss functions include cross-entropy loss \cite{Toutanova2016}, binary negative log likelihood \cite{Trouillon2017}, and max-margin loss \cite{Guu2015,Nickel2016} which we use here:
\begin{align}
\mathcal{L}(\theta) = \sum_{i}^N \sum_{t' \in N(t)} [1 - s_{c_i} + s'_{c_i}]^{+}
\end{align}
$s_{c_i} = s_c(s_i,r_i,t_i)$ and $s^{'}_{c_i} = s_c(s_i,r_i,t^{'}_i)$. $N(t)$ is the set of incorrect targets. Similar triples are used where the relation and target are shared, but the source entity is incorrect.

\subsection{TransE}

TransE \cite{Bordes2013} interprets relations as a translation operation from the source to the target mediated by the relation. More specifically, it embeds a triple spatially such that the source vector can travel to the target vector through the relation vector, i.e., $x_s + x_r \approx x_t$. The scoring function $s_c(s,r,t)$ for TransE is given by
\begin{align*}
s_c(s,r,t) = -d(x_s + x_r - x_t)
\end{align*}
\noindent where $x_s, \ x_r, \ x_t$ are $d$ dimensional vectors, and $d(x)$ is either the $L_1$ or $L_2$-norm of $x$. We use TransE with $L_2$-norm. For learning embeddings, we use max-margin loss (\ref{eqn:loss}). 

Compared to \rescal, TransE has much fewer parameters, but it is more limited in the variety of relations it can model, as the translation operation assumes $1:1$ relations.

\subsection{DistMult}
DistMult \cite{Yang2015} is a special case of the \rescal  model, where the relation matrix is assumed to be diagonal. This results in a sparse relation matrix and consequently fewer parameters. However this simplicity results in the reduction of modeling power. The DistMult model is symmetric and hence can only model symmetric relations. However, DistMult performs well on FB15K benchmark dataset, since the test data contains only a few instances of asymmetric triples. The DistMult scoring function is given by
\begin{align*} 
s_c(s,r,t) = x_s^T \ \textrm{Diag}(W_r) \ x_t
\end{align*}
This can also be written as a three way inner product
\begin{align*} 
s_c(s,r,t) = \langle x_s,x_r, x_t\rangle
\end{align*}
\noindent where $\langle x_s,x_r, x_t\rangle = \sum_i x_{s_i}x_{r_i}x_{t_i}$ and $x_r = \textrm{Diag}(W_r)$ and $x_s,x_r,x_t \ \in \ \mathbb{R}^d$. As before we use the margin loss (\ref{eqn:loss}) for learning these vectors.

\subsection{ComplEx}
The ComplEx model \cite{Trouillon2017} performs sparse tensor factorization of the KG in the complex domain. Nodes and relations are modeled by $d$ dimensional vectors with a real and imaginary part $(Re(x),Im(x))$. This allows ComplEx to model anti-symmetric relations since the three way dot product (inner product) in the complex domain is not symmetric. ComplEx can be seen as DistMult with complex embeddings. The score function of ComplEx is given by:
\begin{align*} 
s_c(s,r,t) &= Re(\langle x_s,x_r, \bar{x_t}\rangle) \\
&=\langle Re(x_s),Re(x_r), Re(x_t)\rangle + \langle Im(x_s),Re(x_r), Im(x_t)\rangle \\
&+\langle Re(x_s),Im(x_r), Im(x_t)\rangle - \langle Im(x_s),Im(x_r), Re(x_t)\rangle
\end{align*}
\cite{Trouillon2017} trained ComplEx with negative log-likelihood. To maintain the same experimental conditions for assessing the efficacy of negative sampling, we train ComplEx with max margin loss (\ref{eqn:loss}).

\section{Negative sampling}\label{sec:negsamp}

Knowledge Graphs capture knowledge as {\it $<$entity, relation, entity$>$} triples, with entities mapped to nodes, and relations to edges. KGs contain only positive instances. While one-class classification solutions have been around for some time \cite{moya93}, for inducing KG embeddings, using negative instances leads to better models. 

Negative instances are not marked in a knowledge graph. The task of link prediction has much in common with other tasks in NLP where (most of) the observed data consists of positive instances. \citep{smith2005} proposed {\it contrastive estimation}, whereby instances that were produced by perturbing the observed ones (and that themselves have not been observed) will serve as negative instances, and the aim is to rank observed instances higher than the unobserved ("negative") ones. In neural probabilistic language models, negative sampling was first proposed in \cite{bengio08} as importance sampling. A sampling solution that was more stable than importance sampling was introduced by \cite{mnih12}, who built upon the noise-contrastive estimation \cite{gutmann12}. In these approaches negative samples are drawn from a non-parametric noise distribution.

For knowledge graphs in particular there are many different ways to produce negative instances based on the graph structure. We present an overview of techniques for producing negative instances from a knowledge graph, and we evaluate their impact on knowledge graph completion, or link prediction. 


\subsection{Random sampling : R}\label{sec:random}

The simplest form of sampling negative instances is to assume a closed world hypothesis and consider any triple that does not appear in the KG as a negative instance. Let 

$K=K^{+} = \{(s_i,r_i,t_i)|y_i=1;i=1,2,\cdots,N\}$  

\noindent denote the complete knowledge graph, where $y_i=1$ represents the presence of a triple $(s_i,r_i,t_i)$ (a positive instance) and $y_i=0$ represents absence. According to the closed world assumption, the set of negatives $K^{-}$ is given by 

$K^{-}=\{(s_i,r_i,t_i)|y_i=0;i=1,2,\cdots,N\}$ 

Since the KG is incomplete this set contains positive triples not present in the KG. Furthermore this set might be very large because the incorrect facts ($O(N^2)$) far outnumber the correct ones.

A simple solution to the scalability problem is randomly sampling a small number of samples from ${K^{-}}$. Given a positive triple $(s,r,t)$ we generate $n_s$ negative triples by sampling $n_s$ target entities from the entity set $\mathcal{E}$. Since the sampling is random, we do not check whether the sampled triples are present in the train and development set, because the probability they are present in $K^{+}$ is negligible. The same procedure is used to generate negative source entities. 

The negatives produced by random sampling may not be very useful: for the positive triple {\it (Tom\_Cruise, starred\_in, Top\_Gun)}, negative targets such as {\it London}  or {\it Mount\_Everest} seem irrelevant. Relevant negative targets should include entities that are movies, such as Terminator, Inception. To obtain such negatives it is necessary to constrain the set of entities from which samples are drawn. We explore such constraints in the following sections. 

\subsection{Corrupting positive instances : C}

We use a method described in \citep{Socher2013} that generates negative instances by corrupting positive instances: for every relation $r$, \citet{Socher2013} collect the sets 

$S = \{s| (s,r,*) \in K^+\}$ and $T = \{t| (*,r,t) \in {K^+}\}$, 

\noindent and produce sets of corrupted triples 

$S' = \{(s',r,t)| s' \in S, (s',r,t) \not \in K^{+}\}$ and 

$T' = \{(s,r,t')| t' \in T, (s,r,t') \not \in {K^+}\}$. 

During training ${K^+}$ consists of triples from training and development set. We sample a number $n_s$ of negative samples from $S'$ and $T'$. Such a method produces negative instances that are closer to the positive ones than those produced through random sampling. 

An issue with this method is that for relations with very few positive instances, there will not be a large enough pool of source and target candidates to corrupt the positive instances. The data analysis shows that this is an issue for the FB15k dataset. For relations where not enough corrupted negative instances can be produced, we supplement this set with randomly produced negative samples. 

\subsection{Typed Sampling : T}

\begin{table*}[h]

\centering
\begin{tabular}{l|lll|l}
Source Type & Source & Relation & Target & Target Type \\ \hline \hline
$film$ & $star\_wars\_episode\_IV$ & $produced
\_by$ & $george\_lucas$ & $film\_producer$ \\
$person$ & $alexandre\_dumas$ & $people\_profession$ & $writer$ & $profession$ \\
$academic\_post$ & $professor$ & $profession\_people$ & $albert\_einstein$ & $award\_winner$ \\ \hline
\end{tabular}
\caption{\textbf{Entity Types in Freebase}: Examples of source and target entity types from Freebase used for generating negative samples.}
\label{table:cats}
\vspace{-0.9cm}
\end{table*}

Knowledge graphs such as FreeBase and NELL \cite{Carlson2010T} have strongly typed relations. For example, a relation $born\_in$ holds between entities of type $person$ and entities of type $city$. Relevant negative candidates (sources or targets) can be mined by constraining the entities to belong to the same type as that of the source (or target). This can help bypass the problem mentioned for the corrupt method, when some relations in the dataset have very few instances. 


For every relation $r : S \rightarrow T$, 

if $S_{r,t} = \{s| s \mbox{ has type } S_t\}$ and $T_{r,t} = \{t| t \mbox{ has type } T_t \}$, 

\noindent with $S_t$ and $R_t$ the domain and range respectively of $r$, negative instances will consist of triples 

$(s',r,t), s' \in S$ and $(s,r,t'), t' \in T$, 

\noindent such that 

$(s',r,t) \not \in R$ and $(s,r,t') \not \in K^{+}$. 

\noindent We then sample $n_s$ number of negative samples from these triples.

If an entity has more than one type (e.g. {\it Albert\_Einstein} has types {\it person, scientist}), we include it in $S_{r,t}$ (or $T_{r,t}$) if one of its types matches $S_t$ (or $T_t$). 
We obtain category data for the Freebase dataset from Freebase relation metadata released in \citep{Gardner2015}, and the entity type by mapping the Freebase entity identifier to the Freebase category. This results in 101,353 instances of the {\it category} relation which is used in the training stage to produce typed negative samples. Domain and range types for Freebase relations are provided by Freebase itself. A few examples of entities and types are included in Table \ref{table:cats}.

We do not use typed sampling for Wordnet. The hypernym/hyponym relations are the de facto {\it type} relations in WordNet, but are hierarchical rather than a mapping onto a given small set of predetermined types as in Freebase.

\subsection{Relational Sampling : REL}

Although typed or corrupt relation sampling can generate relevant negative candidates, due to the incompleteness of the KG, some of these candidates could be unknown positives. If we assume that source target pairs participate in only one relation, then sampling targets (sources) that are connected to the current source (target) through relations other than the current relation can yield true negatives. This is a common procedure in multi-class learning.

More formally, for positive triple $(s,r,t)$ the negative candidate source set is $S^- = \{s|(s,r',t'), \ \forall \ r' \ \in \ \mathcal{R}, r' \neq r\}$ and target set $T^- = \{t|(s',r',t), \ \forall \ r' \ \in \ \mathcal{R}, r' \neq r\}$. As before, after computing S and T we filter out positive triples from train and development set and sample a number $n_s$ of negative samples.

\subsection{Nearest Neighbor sampling : NN}

Most negative sampling methods generate negative samples based on either the closed world assumption, functional constraints such as type constraints, and triple perturbation \cite{Nickel2016b}. We introduce a  negative sampling method which uses a pre-trained embedding model for generating negative samples. We name this pre-trained embedding model the `negative sampling model'. We use the negative sampling model to generate negative targets (sources) that are close to the positive target (source) in vector space. This would help the model learn to discriminate between positives and negatives very similar to the positives.

For a positive triple $(s,r,t)$, with $x_t$ the vector representation of $t$ obtained from the negative sampling model, the set of negative samples are the top $n_s$ nearest neighbors of $x_t$ (that are not positive) obtained from the negative sampling model. The negative sampling model may be different than the model that is being trained. We use the \rescal  model trained with 100 typed (\textbf{T}) negative samples as a negative sampling model for the FB15K dataset. Note that the \rescal  model parameters are frozen (not updated), it is simply used for generating negatives that are used for training another model. Algorithm \ref{alg:nn} describes the procedure for a single triple. In practice we use a batch of triples and the nearest neighbor search is performed using the Ball Tree algorithm which is built only once since the negative sampling model is not updated.
\begin{algorithm} 
\SetKwInOut{Input}{Input}
\SetKwInOut{Output}{Output}
\Input{Triple (s,r,t), Entity Set $\mathcal{E}$, Positive source and targets $P_s$ and $P_t$, Negative Sampling Embedding Model $f_n$, Number of negative samples $n_s$}
\Output{Set of $n_s$ negative samples}
$N_s \leftarrow \mathcal{E} \backslash P_s, \ N_t \leftarrow \mathcal{E}\backslash P_t$\;
$X_{n}^s \leftarrow f(N_s)$, $X_{n}^t \leftarrow f(N_t)$ \;
Initialize the K ball tree with $X_{n}^s$ and $X_{n}^t$ \; 
$x_t \leftarrow \ f_n(t)$ \;
$x_s \leftarrow \ f_n(s)$ \;
$S \leftarrow $ nearest\_neighbors($x_s$,num=$n_s$)\;
$T \leftarrow $ nearest\_neighbors($x_t$,num=$n_s$)\;
\Return S,T
  \caption{Algorithm 1 Nearest Neighbor Sampling}
  \label{alg:nn}
\end{algorithm}
  

Nearest neighbor sampling is computationally expensive compared to the methods discussed in previous sections. This is because a search over all entities needs to be performed for source and target entities for every triple. Therefore we use a model trained using typed negative sampling methods for Freebase and corrupted sampling for Wordnet to initialize the parameters and then fine tune the model using nearest neighbor sampling for 5 epochs.

\subsection{Near Miss sampling : nmiss}
The nearest neighbor sampler generates negatives that are similar to positives in vector space. Some of those negatives may be ranked higher than the positives. Exposing such highly ranked negatives to the classifier can help the model learn a better discriminator. We name this setting as near miss sampling, because the generated negatives are top ranked candidates which makes it difficult for the model to classify them as negatives (near misses). To generate highly ranked negatives, we collect the top $n_s$ targets (sources) closest to the \textit{predicted} target (source) vector. Like the nearest neighbor sampler, we use the negative sampling model for obtaining the predicted vector and entity embeddings. The negative sampling model is not updated.

Given a positive triple $(s,r,t)$ we obtain the predicted vector $v_t = x_s^T \ W_r$ where $x_s, \ W_r$ are entity and relation embeddings of source $s$ and relation $r$ obtained using the negative sampling model. Note that $v_t$ may not be the same as $x_t$, the target entity representation. The set of (target) negative samples are the top $n_s$ nearest neighbors of the predicted vector $v_t$. Algorithm \ref{alg:adv} describes the procedure for a single triple, in practice we use a batch and the Ball Tree is built only once.

\begin{algorithm} 
\SetKwInOut{Input}{Input}
\SetKwInOut{Output}{Output}
\Input{Triple (s,r,t), Entity Set $\mathcal{E}$, Positive source and targets $P_s$ and $P_t$, Negative Sampling Embedding Model $f_n$, Number of negative samples $n_s$}
\Output{Set of $n_s$ negative samples}
$N_s \leftarrow \mathcal{E} \backslash P_s, \ N_t \leftarrow \mathcal{E}\backslash P_t$\;
$X_{n}^s \leftarrow f(N_s)$, $X_{n}^t \leftarrow f(N_t)$ \;
Initialize the K ball tree with $X_{n}^s$ and $X_{n}^t$ \; 
$x_s \leftarrow \ f_n(s)$, $x_t \leftarrow \ f_n(r)$, $W_r \leftarrow \ f_n(r)$ \;
$v_s \leftarrow x_s^T \ W_r$, $v_t \leftarrow W_r \ x_t$ \; 
$S \leftarrow $ nearest\_neighbors($v_s$,num=$n_s$)\;
$T \leftarrow $ nearest\_neighbors($v_t$,num=$n_s$)\;
\Return S,T
  \caption{Near Miss Sampling using \rescal  negative sampler}
  \label{alg:adv}
\end{algorithm}

\begin{figure}
 \includegraphics[scale=0.4]{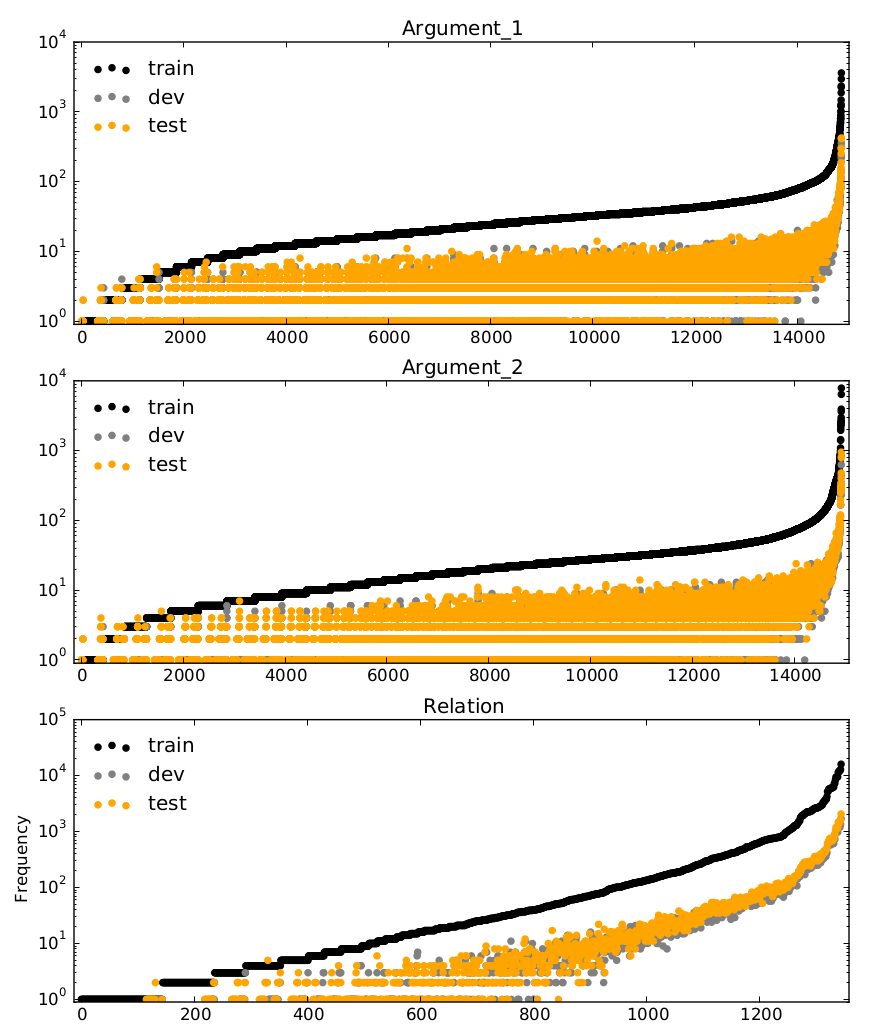}
\vspace{-5mm}
\caption{FB15k dataset frequency statistics}\label{fig:freebase}
\label{fig:fb15k}
\vspace{-5mm}
\end{figure}

Like nearest neighbor sampling, near miss sampling is also computationally expensive, so instead of learning from randomly initialized parameters we tune a pre-trained model for 5 epochs.

\section{Data}\label{sec:data}

\begin{table}[]
\centering
\begin{tabular}{l|lllll}
Data set &  $|\mathcal{E}|$      &  $|\mathcal{R}|$    & Training & Development & Test  \\
\hline
FB15K    & 14,951 & 1345 & 483,142  & 50000       & 59071 \\
WN18     & 40,943 & 18   & 141,442  & 5000        & 5000 
\end{tabular}
\caption{Dataset Details: $|\mathcal{E}|$ = \# of entities,  $|\mathcal{R}|$ = \# of relations.}
\label{table:dataset}
\vspace{-1cm}
\end{table}

We evaluate the impact of negative sampling on the Freebase dataset (FB15k) and on the WordNet dataset (WN18) introduced by \cite{Bordes2013}. They are very different in coverage -- FB15k contains mostly named entities connected through strongly typed relations, while WN18 contains mostly common nouns connected through lexical and semantic relations. Dataset details are included in Table \ref{table:dataset}.

\subsection{FB15k}

FB15k \cite{Bordes2013} consists of approximately 15,000 entities and 1345 relations. We use the split supplied by the dataset: 483,142 train, 50,000 validation and 59,071 positive test instances. 

The training data contains relations that have high variation in the number of instances -- 39\% of the relations have at most 10 instances, while the most frequent relation\footnote{/award/award\_nominee/award\_nominations./award/award\_nomination/award\_nominee} has almost 16000. This disparity is also reflected in the distribution of node degrees -- 12\% of the entities have degree equal or less than 10 (appear in at most 10 instances). The average degree of a node in FB15k is approximately 13.2 overall, and 32.4 on the training data.  The distribution of relations and node degrees is presented in Figure \ref{fig:freebase}.

\begin{figure}
\includegraphics[scale=0.4]{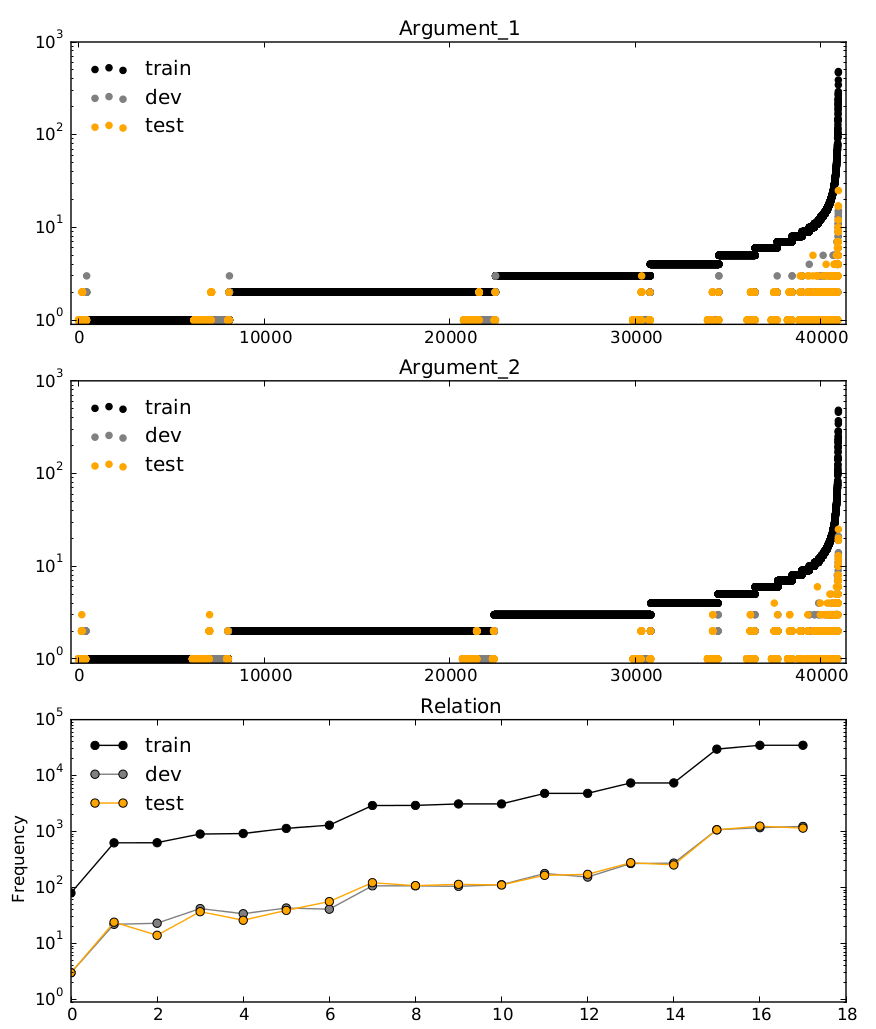}
\vspace{-5mm}
\caption{WordNet18 dataset frequency statistics}\label{fig:wordnet}
\label{fig:wn18}
\vspace{-5mm}
\end{figure}

The type of relations included in Freebase connect named entities. They are extrinsic relations, in that they do not hold based on the intrinsic properties of the connected entities, but are due to external circumstances. For example, the {\it people\_profession} relation connecting people and their professions are not determined by intrinsic properties of people and professions. Relations in FreeBase are strongly typed -- the domain and range of the relations are types, e.g. the {\it country\_capital} relation connects {\it countries} and {\it cities}.

\subsection{WN18}

This dataset consists of a subset of relations from the WordNet lexical database\footnote{\url{https://wordnet.princeton.edu/}}, split into training, development and testing: 141442/ 5000/ 5000. There are 18 relations. There is less variation in the number of instances per relation compared to the FB15k, as can be seen in Figure \ref{fig:wordnet}. There is one relation with less than 100 instances ({\it similar\_to}), while the most frequent relations ({\it hypernym, hyponym}) have approximately 35,000. 

From a graph structure point of view, WN18 nodes have low connectivity -- the average degree on the entire dataset is approximately 1.2, and on the training data alone approximately 3.45. This translates into sparser adjacency matrices for factorization, compared to Freebase.

WordNet contains lexical and semantic relations. Lexical relations -- such as {\it derivationally\_related\_form} connect lemmas from different parts of speech that are morphologically connected. The semantic relations cover {\it is\_a} relations (hypernym / hyponym, instance hypernym/hyponym), three types of {\it part\_of} relations (member, substance and part). The semantic relations in WordNet are intrinsic, as they reflect or arise from intrinsic properties of the connected entities. For example, a cat {\it is\_a} animal, and cat {\it has\_part} paws not because of external circumstances, but because of what a cat is. Compared to FreeBase, WordNet relations are not typed -- there is no clear domain and range for the WordNet relations.

\section{Experiments}

\subsection{Implementation}

For fair comparison we reimplemented \rescal, TransE, DistMult, ComplEx using PyTorch, and tested them using the same experimental setting: same loss (max-margin loss), embedding size (100), and data. We use the Adam \cite{Kingma2014} SGD optimizer for training because it addresses the problem of decreasing learning rate in AdaGrad. We ensure that entity embeddings for all the models have unit norm. We performed exhaustive randomized grid search \cite{bergstra2012} for the $L_2$ regularizer on the validation set for all models and we tuned the training duration using early stopping. The learning rate ($lr$) and $\lambda$ (the $L_2$ norm coefficient) are presented in Table \ref{tab:params}. The code is available in Github \footnote{https://github.com/bhushank/kge-rl}.

\begin{table}
\begin{tabular}{lrr}
Model & $lr$ & $\lambda$ \\ \hline \hline 
\multicolumn{3}{l}{Freebase} \\ \hline		
ComplEx	& 0.001 & 1.31E-06 \\
DistMult & 0.001 &	4.93E-06 \\
\rescal	& 0.001	& 0.0002084 \\	
TransE	& 0.001	& 0.00024036 \\	 \hline \hline
			
\multicolumn{3}{l}{Wordnet} \\ \hline	
ComplEx ($n_s \in \{1,2,5\}$) &	0.005 &	2.82E-05	\\
ComplEx ($n_s  >= 10$)	& 0.01	& 2.82E-05	\\
DistMult ($n_s \in \{1,2,5\}$) &	0.005 &	3.12E-06	\\
DistMult ($n_s  >= 10$) &	0.01 &	3.12E-06	\\
\rescal ($n_s \in \{1,2,5\}$) & 0.005 & 7.48E-05 \\
\rescal ($n_s  >= 10$) & 0.01	& 7.48E-05	\\
TransE ($n_s \in \{1,2,5\}$) &	0.005 &	0.0001863777692	\\
TransE ($n_s  >= 10$) &	0.01 & 0.0001863777692	\\ \hline
\end{tabular}
\caption{Parameter values}\label{tab:params}
\vspace{-1cm}
\end{table}

The different methods for negative sampling described in Section \ref{sec:negsamp} were used to produce negative instances for training. In FB15K some relations do not have enough sources or targets to generate negative triples by corrupting positive triples. If the number of generated triples are less than the required ($n_s$), we complete the set of negative samples with randomly generated triples. 

For the nearest neighbor and near miss settings, we used the best performing model for initializing the parameters, and used the \rescal  model tuned on typed negative samples (100 negative samples) as the negative sampling model for FB15K and \rescal  trained by corrupting positive samples (100 negative samples) for WN18.

\subsection{Test data}

The test data is the same across all experiments. The negative instances for the test data were generated as described in \cite{Bordes2013} -- corrupting positive instances using all entities of the dictionary instead of the correct source and target, without sampling.

Also following the procedure of \cite{Bordes2013}, we use the filtered setting: the negative samples added to the training data are filtered with respect to the test data to avoid (known) false negatives in training. 

\subsection{Evaluation metrics}

For evaluation we use the mean reciprocal rank (MRR) and hits@K that are commonly used for link prediction. 

For a list of N answers for link prediction, the mean reciprocal rank (MRR) and hits@k are defined as: 

\begin{tabular}{p{3.5cm}p{3.5cm}} \\
$MRR = \frac{1}{N} \sum\limits_{i=1}^N \frac{1}{rank_i}$ & $hits@K = \frac{|\{i| rank_i < K\}|}{N} $ \\
\end{tabular}

\noindent where $rank_i$ is the rank of the positive instance $i$ predicted by the model with respect to the negative samples. For FB15k we use hits@10, for WN18, hits@1.

\subsection{Results}

\begin{figure*}
\includegraphics[height=4.2cm,width=0.92\textwidth]{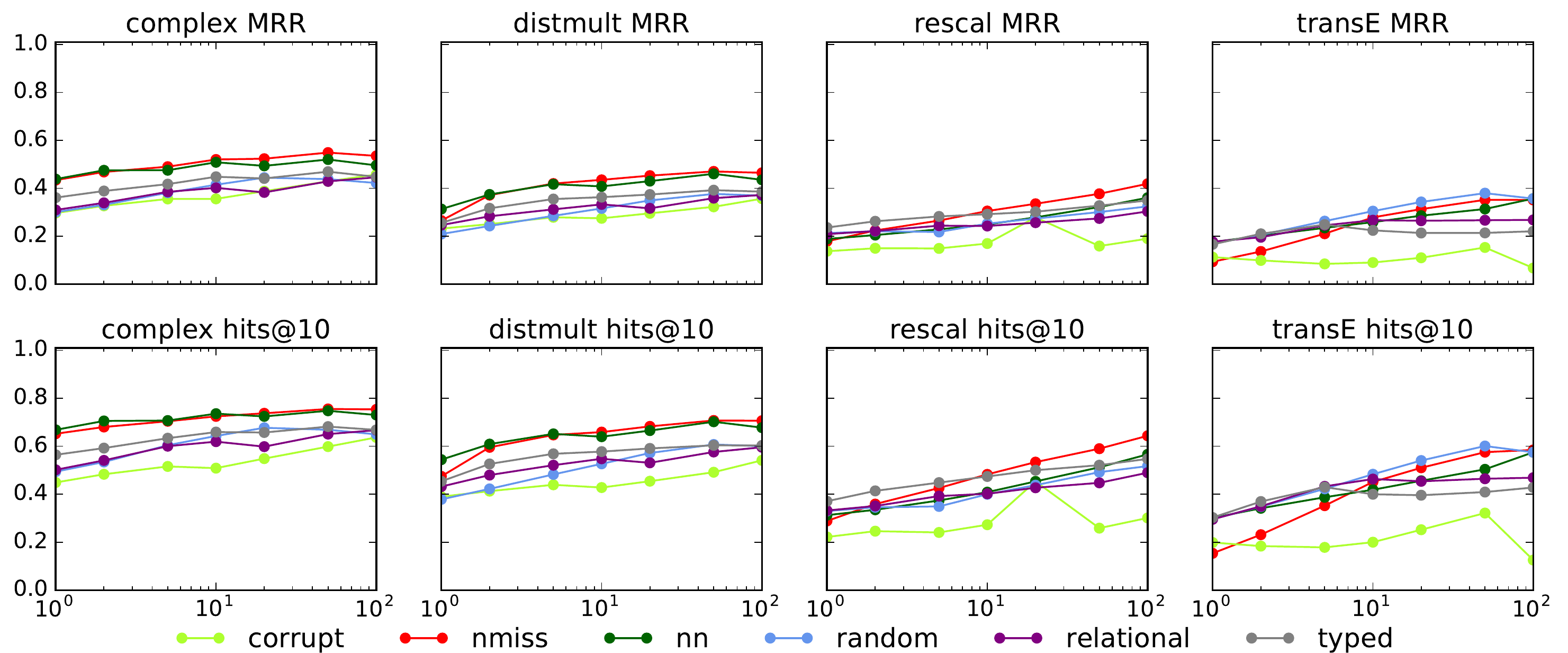}
\vspace{-3mm}
\caption{Link prediction on FB15k, evaluated in terms of MRR
for $n_s \in  \{1, 2, 5, 10, 20, 50, 100\}$ on a logarithmic scale.}\label{fig:freebase_linkpred}
\vspace{-3mm}
\end{figure*}

\begin{figure*}
\includegraphics[height=4.2cm,width=0.92\textwidth]{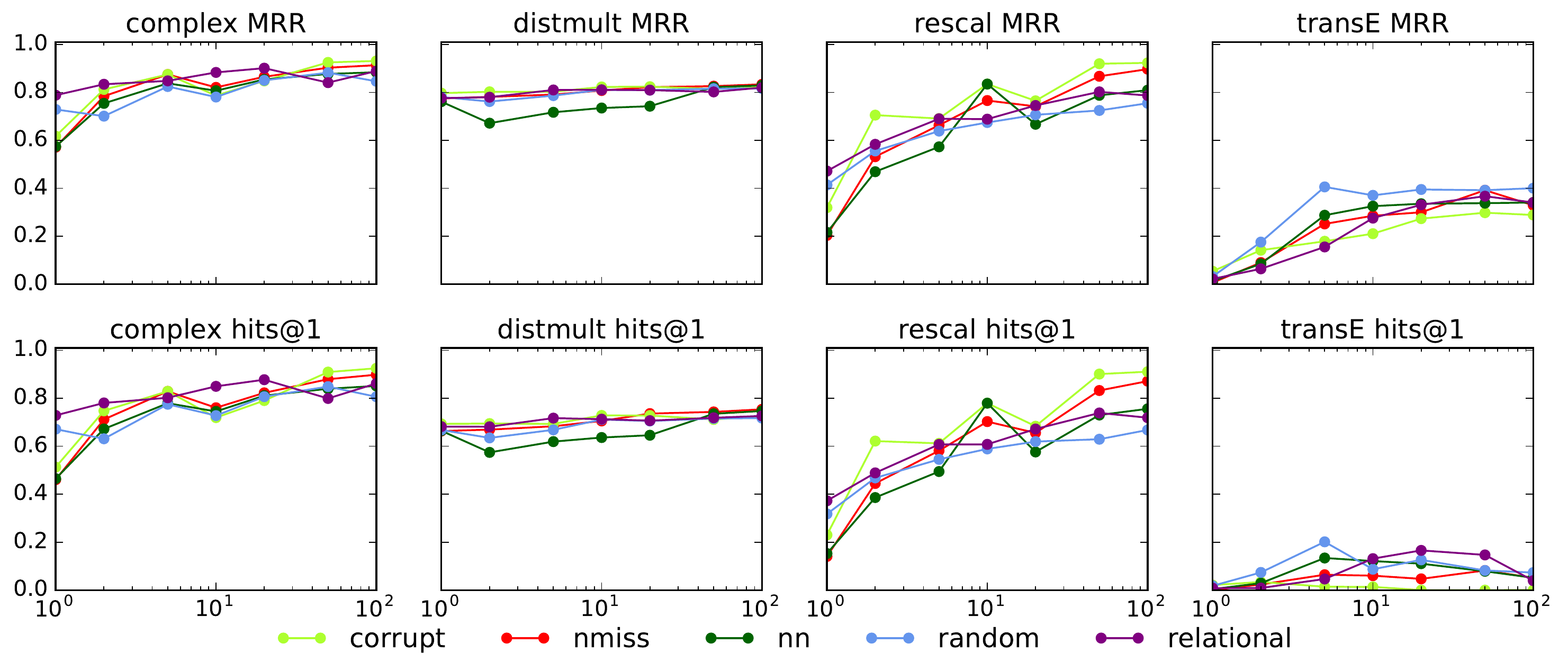}
\vspace{-3mm}
\caption{Link prediction on WN18, evaluated in terms of MRR
for $n_s \in  \{1, 2, 5, 10, 20, 50, 100\}$ on a logarithmic scale.}
\label{fig:wordnet_linkpred}
\vspace{-3mm}
\end{figure*}

We present the results of link prediction on FB15k and WN18 in terms of MRR in Figures \ref{fig:freebase_linkpred} and \ref{fig:wordnet_linkpred} for $n_s \in \{1, 2, 5, 10, 20, 50, 100\}$ for each positive instance. 

The results show that the different sampling methods have different effects on the two datasets. Since link prediction is based exclusively on the embedding of the graphs, differences in performance are caused by the different structure (e.g. different node degrees which are reflected in the sparsity of the relation adjacency matrices) and the different nature of the relations -- typed and extrinsic in FB15k, not typed and (mostly) intrinsic in WordNet. 

As suggested by work on learning statistical models through noise contrastive estimation \cite{Gutmann2010}, selecting difficult negative instances produces better models: near miss sampling leads to better results on FB15k for most embeddings methods. The reason embedding based sampling works well on FreeBase is primarily because the negative samples generated by the pre-trained embedding model are very close to the discriminator boundary. For example, the near miss sampling involves generating negative target entities that are highly ranked by the embedding model. These entities are likely to be highly ranked by the model that is being trained.  Therefore providing these entities as negatives allows the system to learn a model that ranks them below the positive target using the max-margin loss. Note that the samples generated by the embedding model are close to each other in vector space due to the ability of the embedding model to cluster entities. Therefore almost all the generated negative samples are close to the discriminator boundary. We treated the negative sampling model (pre-trained model) as a hyper parameter. We found that the RESCAL model worked best. We speculate that this might be due to the superior ability of RESCAL model to cluster similar entities.

Corrupting positive instances, the method most frequently used for link prediction, is the least competitive on FB15k, but fits WordNet well, particularly for \rescal. DistMult is not very sensitive to the type of negative sampling on WN18, except for the nearest neighbor method with which it does not perform well. 


To understand why corrupting positive instances works best on WordNet, we look at the data and the graph statistics. The WN18 dataset has 18 relations while with FB15k has about 1495 relations. Due to per relation data sparsity in FB15K, see Fig. \ref{fig:fb15k} and \ref{fig:wn18}, negative sampling using corrupted triples works poorly for FB15K, as it often has to fall back on random sampling when not enough positive instances with a shared source/target are available for "corruption". Corrupt sampling works better in an instance rich environment.

Apart from data sparsity, the nature of WordNet and Freebase relations may also affect the performance of negative sampling methods. WordNet relations have open ended ranges and domains while Freebase relations have typed ranges and domains. Embedding based methods, such as the near miss sampling method we implemented, work on the basis of clustering similar entities, and do not function well for WordNet where the relations do not have domains and ranges that reflect conceptual/semantic clusters.

We have discussed the differences in performance of sampling methods for the two KGs used. There are also differences with respect to the link prediction methods. Random sampling works best for TransE. This may be surprising at first, but is understandable considering that the theoretical model behind TransE assumes $1:1$ relations. Providing it with negative entities that are close (using typed, corrupted or embedding methods) does not result in improvement because the negative entities generated using typed, corrupt or embeddings are close to each other in vector space and the model will ultimately be unable to distinguish between them. This is not the case when doing random sampling, when TransE is not perturbed by too close negatives. ComplEx and DistMult perform well with both near miss and nearest neighbour sampling on FB15k. \rescal performs best with near miss sampling on this data, and with corrupting positive samples for WordNet. For middle-range $n_s$ relational sampling performs best.

\begin{figure}
\hbox{\hspace{-0.5cm} \includegraphics[scale=0.65]{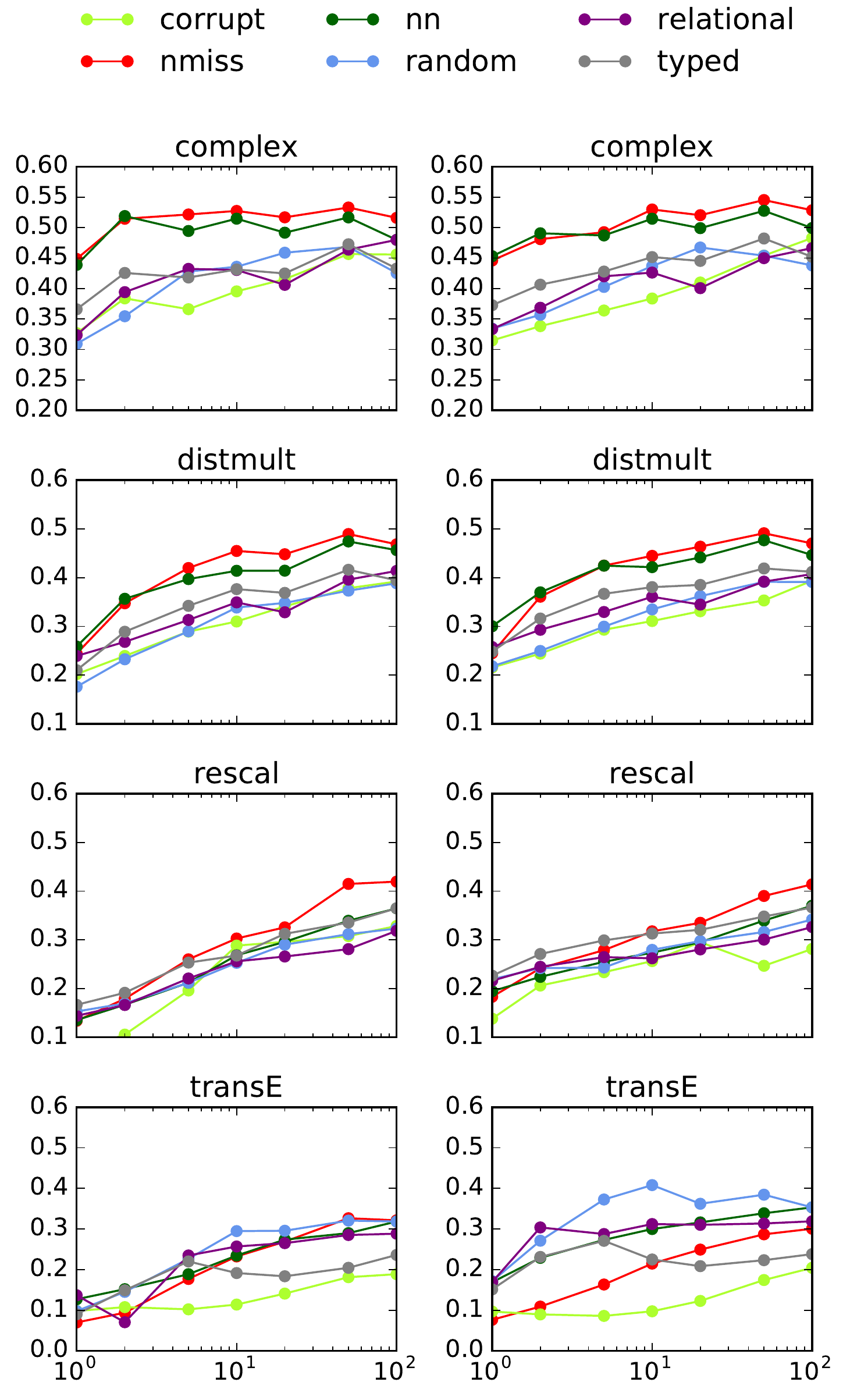}}
\vspace{-4mm}
\caption{Results on relations with OOM 0 and 1 in FB15k (MRRs)} \label{fig:freebase_details}
\vspace{-5mm}
\end{figure}

\begin{figure}
\hbox{\hspace{-0.5cm} \includegraphics[scale=0.65]{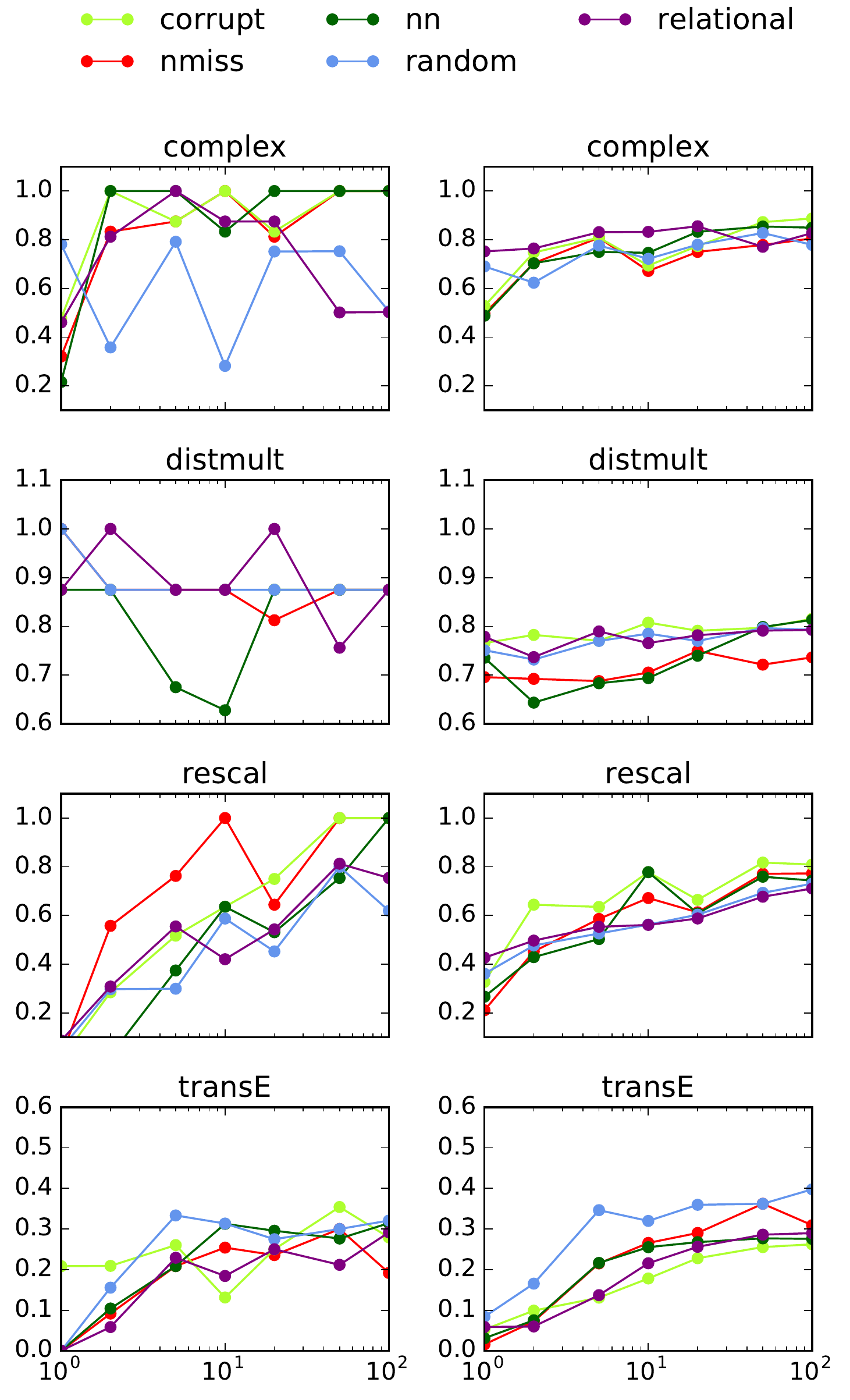}}
\vspace{-4mm}
\caption{Results on relations with OOM 1 and 2 in WN18 (MRRs)} \label{fig:wordnet_details}
\vspace{-5mm}
\end{figure}

As described in Section \ref{sec:data}, the training data for both methods varies quite a bit in terms of the frequency of the relations covered. Freebase is more extreme, in that approximately 39\% of the relations have at most 10 positive instances to train on. We analyzed the effects of negative sampling on different slices of the data, split by the order of magnitude (oom) of the frequency of the relations in the training data. More precisely, we group relations into sets $G_n$ indexed by the order of magnitude $n$:

$G_n = \{ r| 10^{n} < freq(n,\mbox{training data}) <= 10^{(n+1)}\}$\footnote{We include relations that have only one instance in $G_0$.}. 

Freebase has 5 slices (0..4) and WordNet 4 (1..4). The results (as MRR and hits@K) for slices representing relations with OOM 2 or more closely mirror the overall results. The results for the low frequency relations are shown in Figures \ref{fig:freebase_details} and \ref{fig:wordnet_details}.
The hits@K score are similar to the MRR ones, so we do not include them\footnote{The complete set of plots accompanies the code and will be shared.}. 

\begin{table}
\begin{tabular}{l|cc|lcc}
         & \multicolumn{2}{l}{\citet{Yang2015}} & \multicolumn{3}{l}{Negative sampling} \\ \hline
         & MRR     & HITS@10        &  neg. sampling & MRR         & HITS@10 \\ \hline
\multicolumn{5}{l}{FB15k} \\ \hline          
DistMult & 0.35    &  57.7          & near miss    & {\bf 0.46}      & {\bf 70.64}   \\
\rescal  & 0.31    &  51.9          & near miss    & {\bf 0.42}      & {\bf 64.34}   \\
TransE   & 0.32    &  53.9          & near miss    & {\bf 0.37}      & {\bf 62.97}   \\ \hline
\multicolumn{5}{l}{WN18} \\ \hline
DistMult &  {\bf 0.83}   &  {\bf 94.2}          & corrupt    & 0.82        &  94.06   \\
\rescal  &  0.89   &  92.8          & corrupt    & {\bf 0.92}        &  {\bf 93.91}   \\
TransE   &  0.38   &  {\bf 90.9}          & corrupt    & {\bf 0.40}        &  86.98   \\ \hline

\end{tabular}
\caption{SotA results 
using a max-margin loss function and corrupting positive instances vs. the best performing negative sampling.}\label{tab:results}
\vspace{-1cm}
\end{table}

While the results on the low frequency relations cannot be analyzed separately from the other relations because the embeddings process relies on processing and inducing jointly all relation and entity representations, we can note that the performance on link prediction for these relations with very few instances varies much with the negative sampling method. Overall, the best results are obtained with the same sampling method as for their more populous counterparts, but for specific ranges of the number of generated negative samples other methods would work best (e.g. nearest neighbor and relational sampling for WordNet data).

The reported experiments were performed using the max margin loss function. In Table \ref{tab:results} we include the state of the art results on DistMult, \rescal and TransE obtained with a max margin loss function reported in \cite{Yang2015} and corrupting tripes, to compare with the results obtained with the best negative sampling method for the dataset. Slight differences in the learning rate and $\lambda$ account for the differences in performance when using corrupt positive instances as negative samples for the WN18 dataset.

Recently, \cite{Trouillon2017} used the log-likelihood objective, which leads to improvements over the published results for the methods they compared (TransE, ComplEx, HolE, DistMult). We plan to analyze the negative sampling methods while using this new loss function.


\section{Conclusion}

We report an analysis of the impact of six negative sampling methods on the performance of link prediction in knowledge graphs, for four methods for graph embedding -- ComplEx, DistMult, \rescal, TransE. The analysis is performed with respect to two datasets -- a subset of Freebase (FB15k) and a subset of WordNet (WN18) -- that are very different in the type of knowledge they cover.

The results indicate that different approaches to negative sampling work best for the two resources. The proposed near miss sampling worked best for Freebase with most of the graph embedding methods, while corrupting positive triples leads to best results on WordNet. The newly proposed near miss and nearest neighbor negative sampling work best for Freebase, for three out of the four graph embeddings methods. From analysis of datasets, we further concluded that embedding based negative sampling is very useful for combating data sparsity, while corrupt sampling works best in the data rich scenario. The nature of the relations in these graphs (typed with respect to their domain and range vs. open) as well as the statistics of the knowledge graph (number of positive instances per relation) explain the different behaviour with respect to negative sampling.

\bibliographystyle{ACM-Reference-Format}
\bibliography{kbcom} 

\onecolumn
\appendix
~\\
\paragraph{Performance analysis on Freebase (FB15k)} ~\\

\begin{figure}[h]
\includegraphics[scale=0.6]{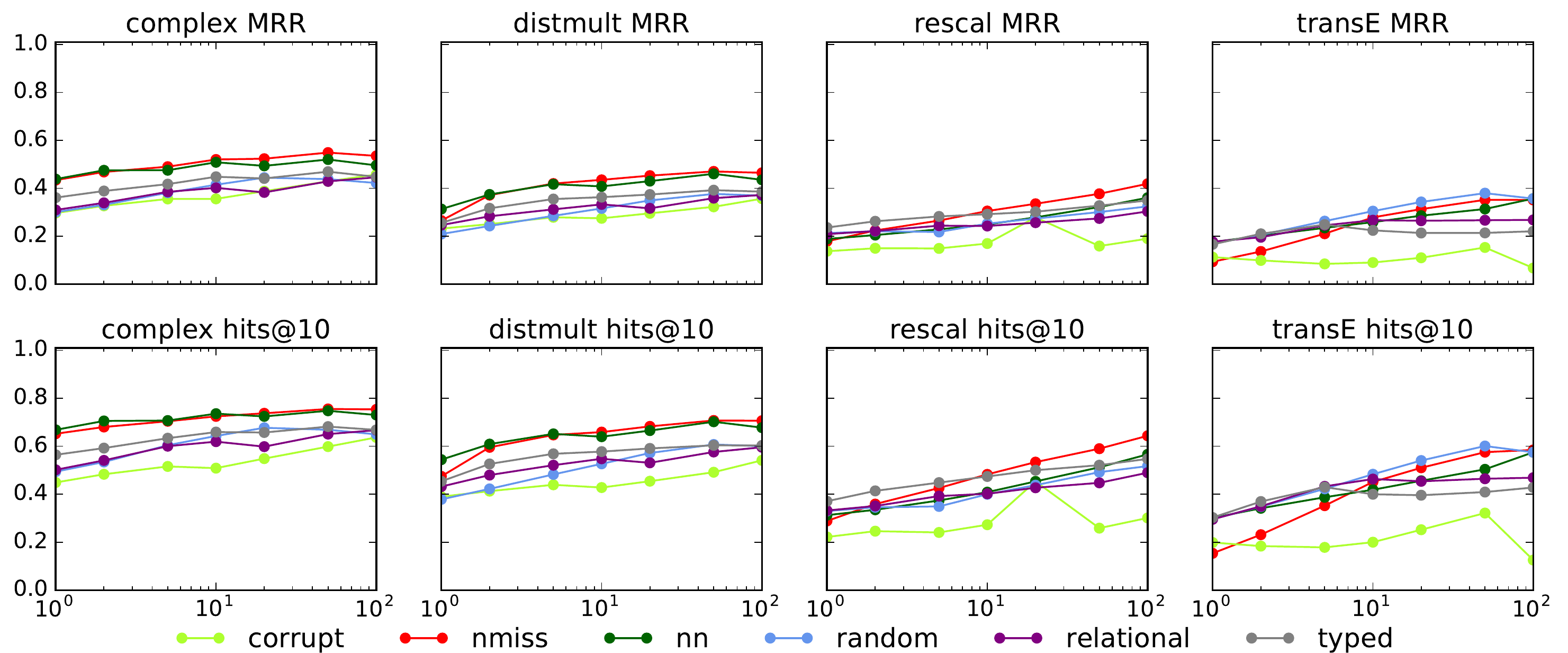}
\caption{Performance on FB15k in terms of MRR and Hits@10}
\end{figure}

~\\
\paragraph{Performance analysis on WordNet (WN18)} ~\\

\begin{figure}[h]
\includegraphics[scale=0.6]{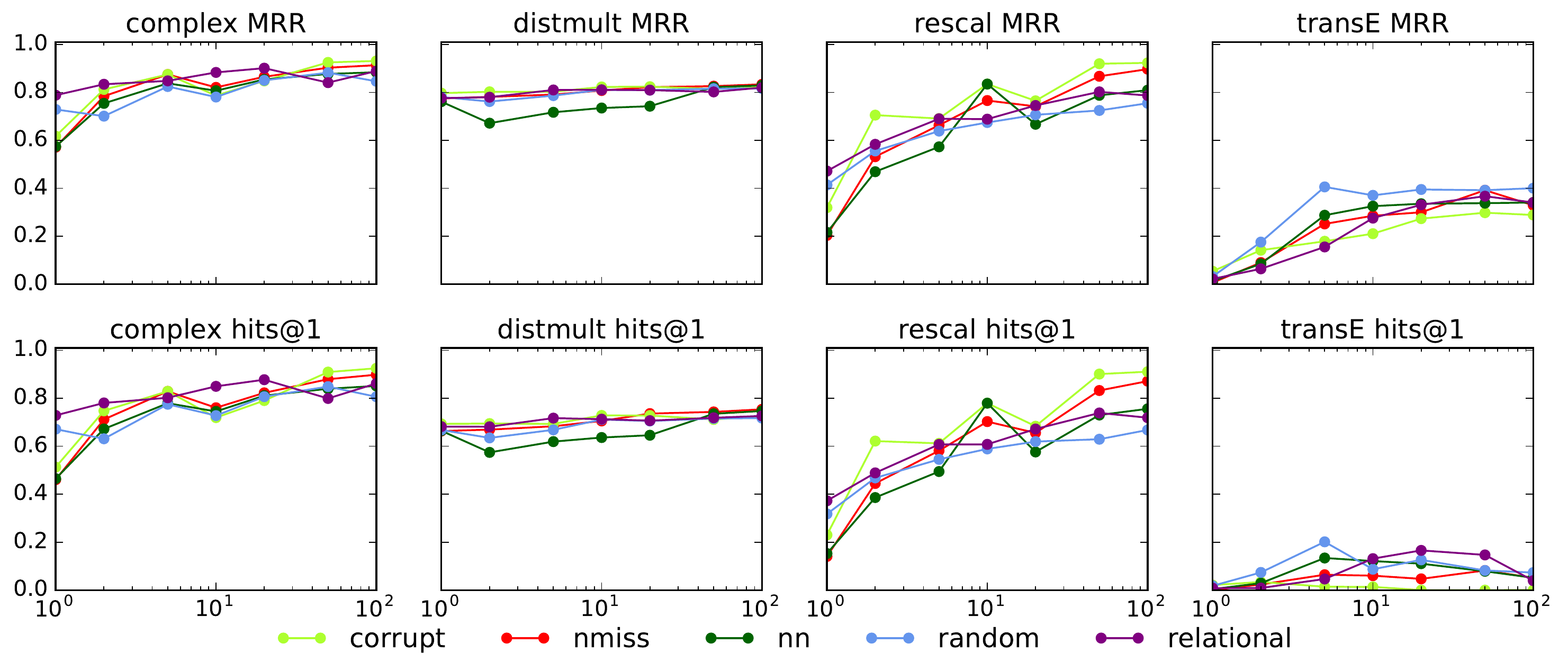}
\caption{Performance on WN18 in terms of MRR and Hits@1}
\end{figure}

\onecolumn
\begin{figure}
\rotatebox{90}{
	\includegraphics[scale=0.5]{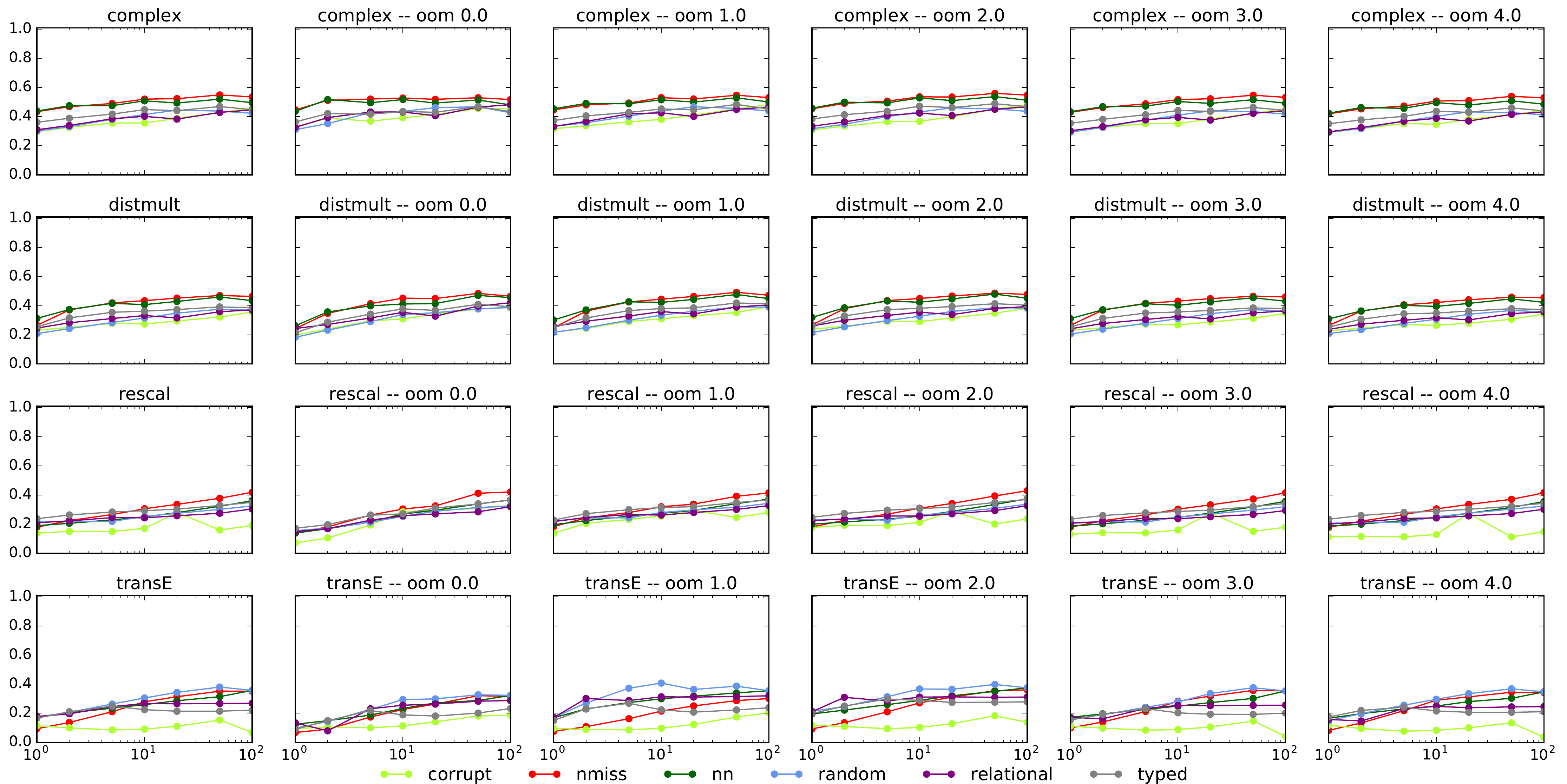}
    }
\caption{Performance on FB15k in terms of MRR for relations with different orders of magnitude}
\end{figure}

\begin{figure}
  \rotatebox{90}{
	\includegraphics[scale=0.5]
{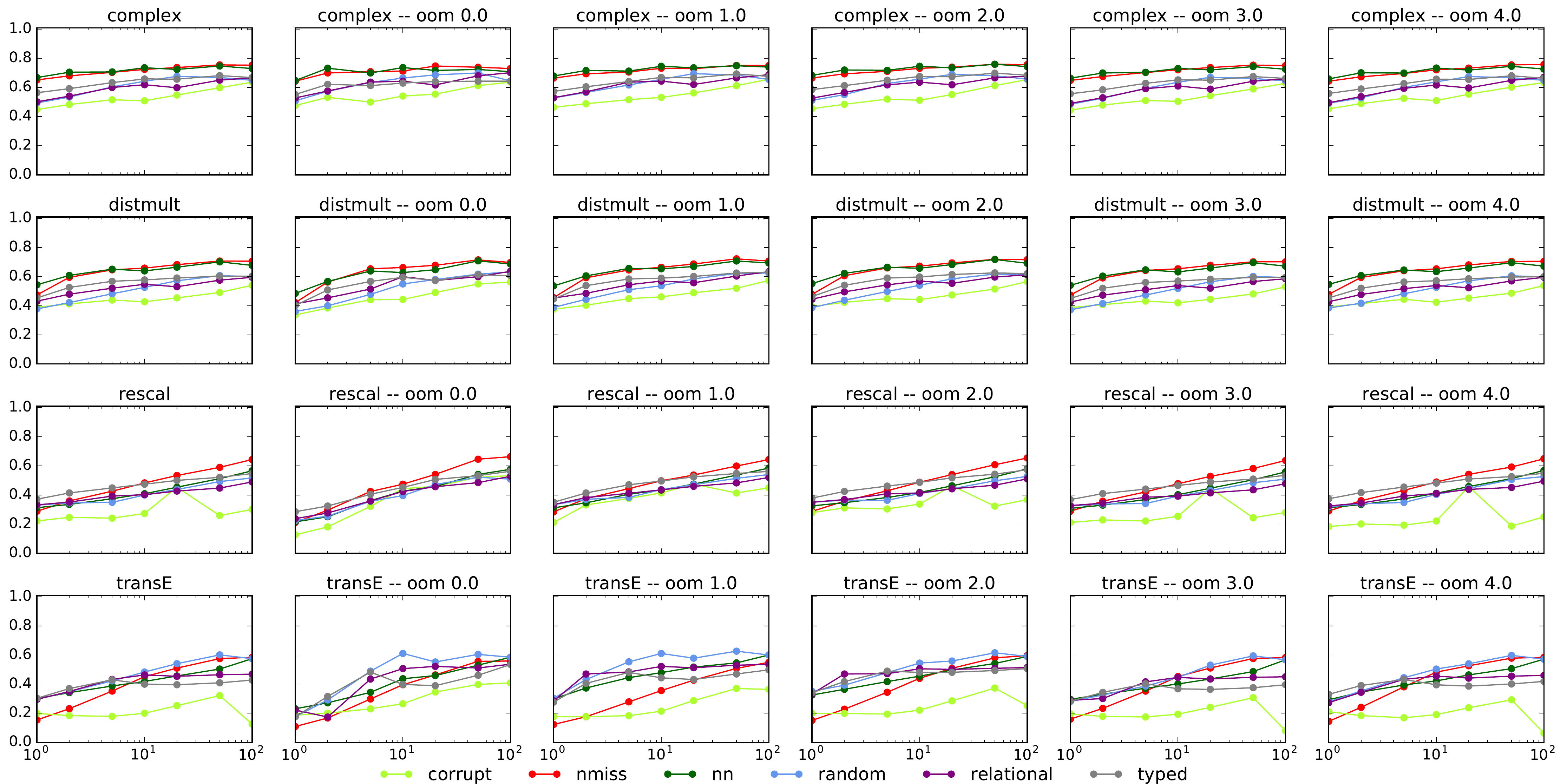}
}	
\caption{Performance on FB15k in terms of Hits@10 for relations with different orders of magnitude}
\end{figure}

\begin{figure}
  \rotatebox{90}{
	\includegraphics[scale=0.6]{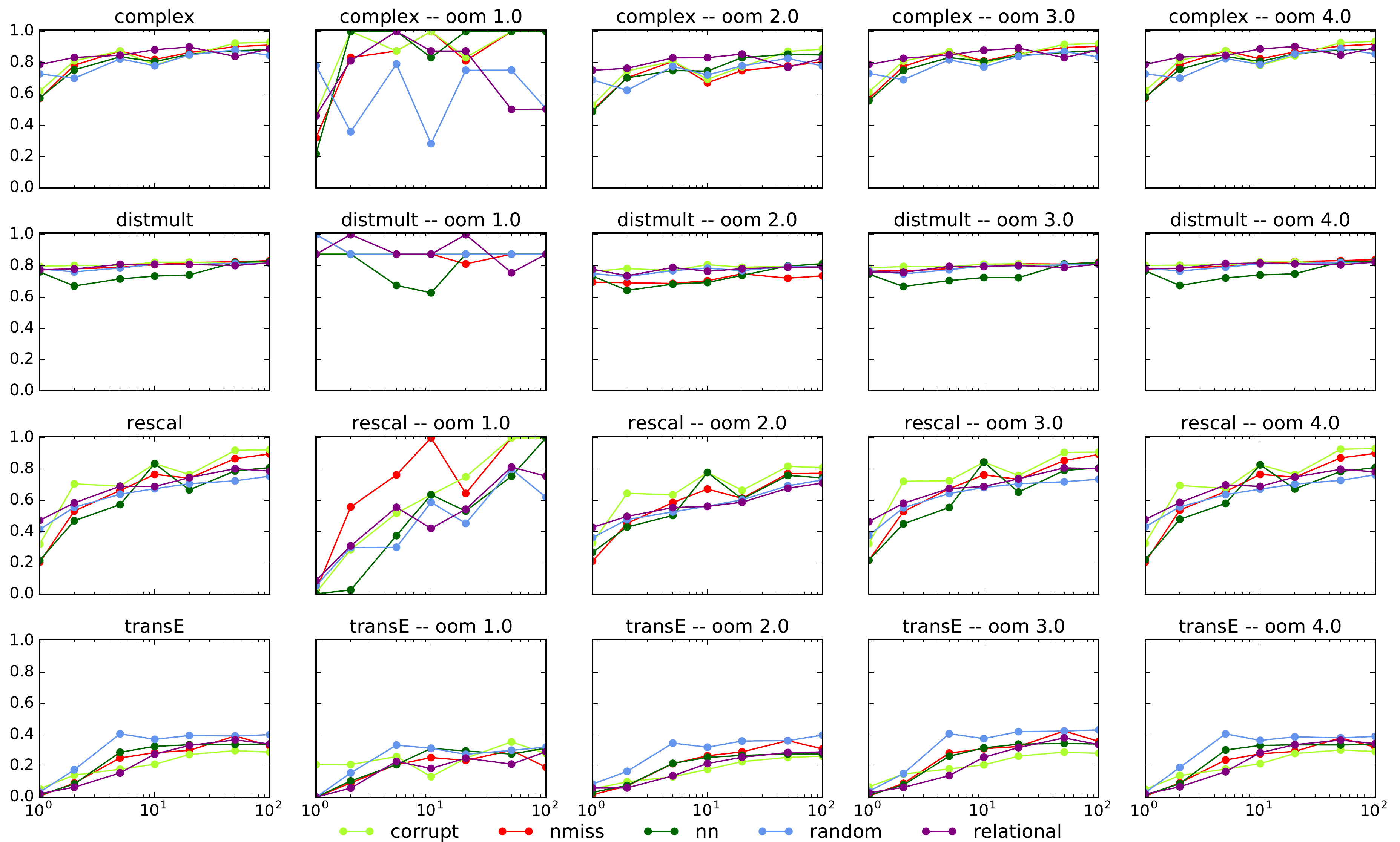}
    }	
\caption{Performance on WN18 in terms of MRR for relations with different orders of magnitude}
\end{figure}

\begin{figure}
\rotatebox{90}{
	\includegraphics[scale=0.6]{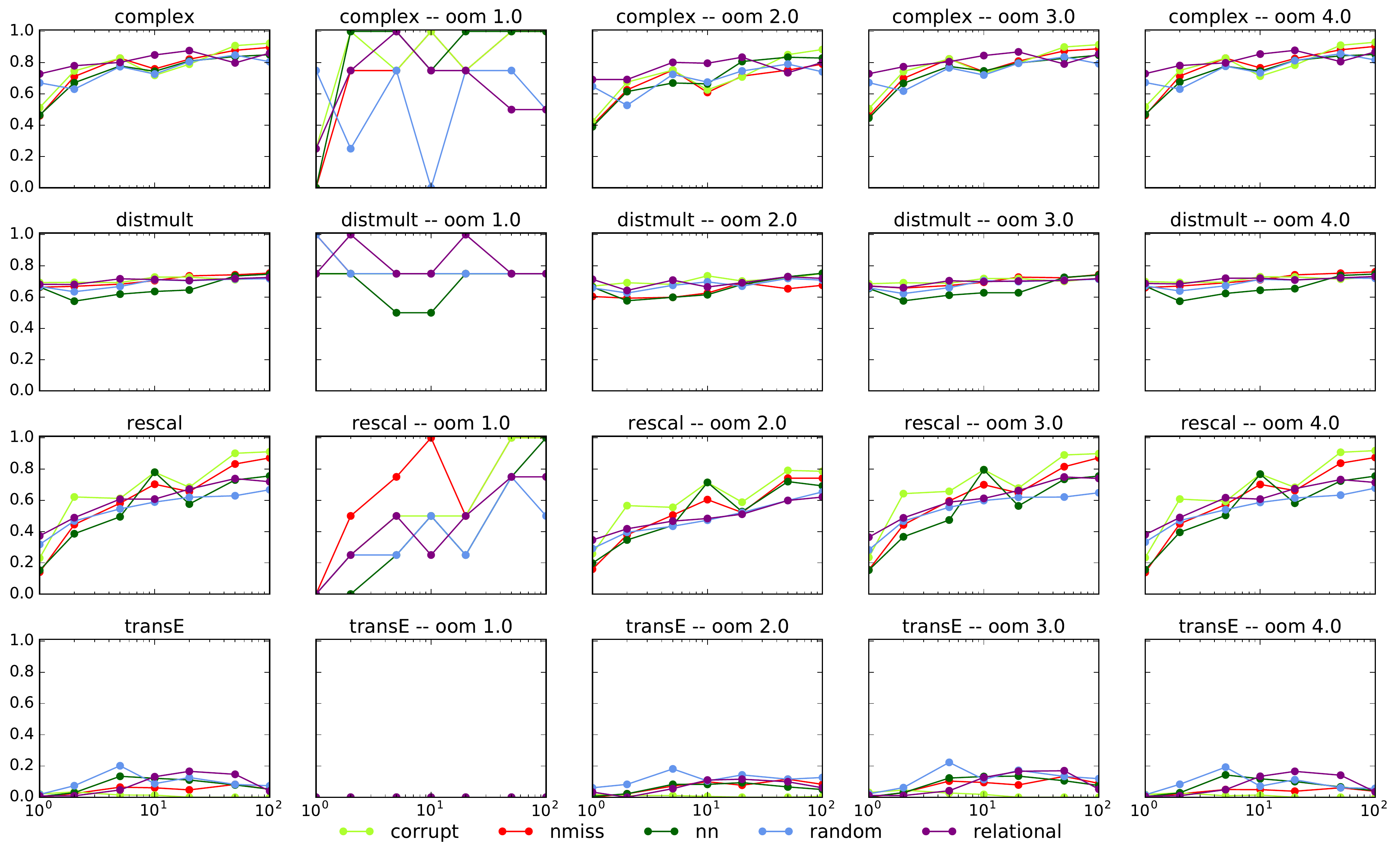}
    }	
\caption{Performance on WN18 in terms of Hits@1 for relations with different orders of magnitude}
\end{figure}

\end{document}